\documentclass[10pt,journal,final,twocolumn]{IEEEtran}

\usepackage{graphicx,times,amsmath,amssymb,cite,cases,mathrsfs,booktabs,multirow,psfrag,subfigure,url,stfloats,threeparttable}
\usepackage{algorithm,algorithmic,amsthm,ulem,latexsym,bm}
\usepackage{dsfont}
\usepackage[colorlinks,linkcolor=red,anchorcolor=blue,citecolor=red]{hyperref}

\begin{document}

\title{Dual Adversarial Auto-Encoders for Clustering}
\author{Pengfei Ge, Chuan-Xian Ren, Jiashi Feng, Shuicheng Yan,~\IEEEmembership{Fellow,~IEEE}% <-this % stops a space
\thanks{P.F. Ge and C.X. Ren are with the Intelligent Data Center, School of Mathematics, Sun Yat-Sen University, Guangzhou, China. They contributed equally to this work. C.X. Ren is the corresponding author.}
\thanks{J.S. Feng and S.C. Yan are with the Department of Electrical and Computer Engineering, National University of Singapore.}
\thanks{This work is supported in part by the National Natural Science Foundation of China under Grant 61572536, in part by the Science and Technology program of Guangzhou under Grant 201804010248.}}

\date{}
%\maketitle
\IEEEcompsoctitleabstractindextext{%

\begin{abstract}
As a powerful approach for exploratory data analysis, unsupervised clustering is a fundamental task in computer vision and pattern recognition. Many clustering algorithms have been developed, but most of them perform unsatisfactorily on the data with complex structures. Recently, Adversarial Auto-Encoder (AAE) shows effectiveness on tackling such data by combining Auto-Encoder (AE) and adversarial training, but it cannot effectively extract classification information from the unlabeled data. In this work, we propose Dual Adversarial Auto-encoder (Dual-AAE) which simultaneously maximizes the likelihood function and mutual information between observed examples and a subset of latent variables. By performing variational inference on the objective function of Dual-AAE, we derive a new reconstruction loss which can be optimized by training a pair of Auto-encoders. Moreover, to avoid mode collapse, we introduce the clustering regularization term for the category variable. Experiments on four benchmarks show that Dual-AAE achieves superior performance over state-of-the-art clustering methods. Besides, by adding a reject option, the clustering accuracy of Dual-AAE can reach that of supervised CNN algorithms. Dual-AAE can also be used for disentangling style and content of images without using supervised information.
\end{abstract}

}

\maketitle \IEEEdisplaynotcompsoctitleabstractindextext \IEEEpeerreviewmaketitle

\section{Introduction}

\IEEEPARstart{A}{s} a fundamental task in pattern recognition and machine learning, clustering \cite{jain1999data,Frey2007clustering,David2017cluster,Ren2019pr} is an effective approach for exploring the structure of unlabeled data and extracting the classification information within. However, it is quite challenging since classification information is always entangled with other information such as style and background.

To date, various clustering methods have been proposed, and most of them measure the similarity between samples for clustering thus fail to capture a disentangled representation. Some popular ones include $K$-means \cite{macqueen1967some}, Gaussian Mixture Models (GMM) \cite{hastie2009overview} and Spectral clustering \cite{ng2002spectral}. Recently, deep generative models achieve noticeable performance, which learn the distribution of data under some assumptions w.r.t. the latent variables and thus encode rich latent structures. Variational Auto-Encoder (VAE)~\cite{kingma2013auto} is an important framework for training generative models, in which the Stochastic Gradient Variational Bayes and re-parametrization trick are utilized to maximize the likelihood function. However, a VAE-based clustering model would suffer over-regularization \cite{kingma2016improving} thus clustering degeneracy. Adversarial Auto-Encoder (AAE)~\cite{makhzani2015adversarial} combines AE with an adversarial training mechanism. It uses an adversarial training mechanism to match the aggregated posterior distribution of the hidden code vector with an arbitrary prior distribution, which enables the model to pursue complex prior distribution more closely. AAE can achieve good performance in many applications such as disentangling style and content of images, dimensionality reduction and data visualization. However, it is still challenged by the following issues. First, the objective of AAE is to maximize the likelihood function, which does not accord with the objective of clustering tasks; second, due to the adversarial training mechanism, the model tends to suffer mode collapse\footnote{Mode collapse means samples generated by the generator lack diversity.}.

\begin{figure}[tb]
\centering{\scalebox{0.47}{\includegraphics{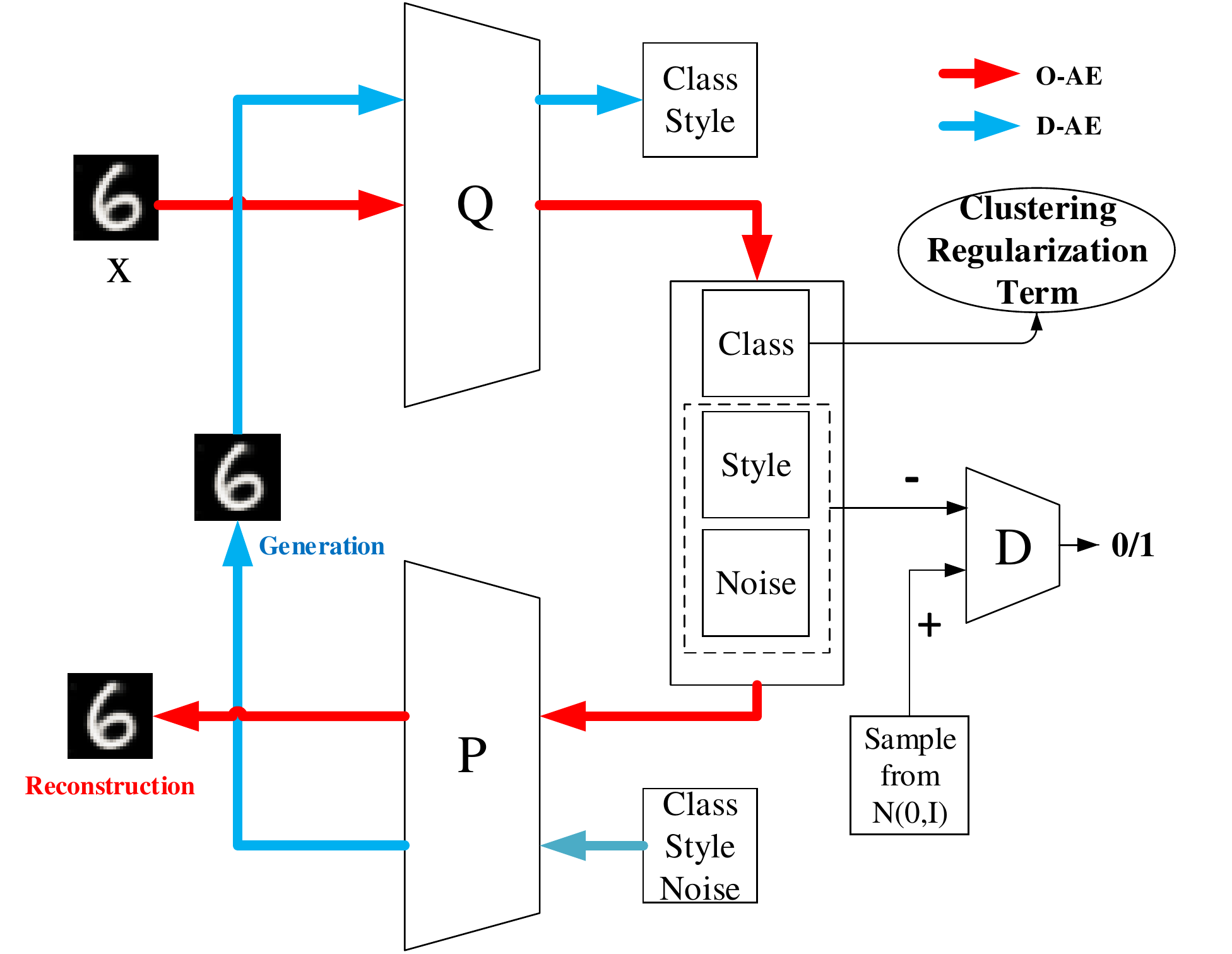}}\caption{The architecture of Dual-AAE. Dual-AAE consists of two AEs: O-AE and D-AE. O-AE (red lines) reconstructs the input data from the category, style and noise variables. The GAN and the clustering regularization term are used for regularizing the hidden code. D-AE (green lines) is used to generate new samples and reconstruct the class variable and the style variable. Better viewed in color.}\label{fig0}}
\end{figure}

In this paper, we propose a novel deep generative model for clustering, named Dual Adversarial Auto-encoder (Dual-AAE), which can effectively extract classification information and will not be threatened by model collapse. In particular, besides maximizing the likelihood function of observed samples as in VAE and AAE, the proposed Dual-AAE also maximizes the mutual information between observed samples and a subset of their latent variables in order to capture more structure information contained in the data. Such a mutual information term cannot be computed directly but we can derive its lower bound via variational inference. Note in the Dual-AAE, the same variational auxiliary distribution is used to obtain the variational lower bound of both the likelihood function and the mutual information term.
Correspondingly, it consists of a pair of Auto-encoders (AE), namely O-AE and D-AE, respectively for reconstructing the input data and reconstructing the latent variables. These two auto-encoders share the same network parameters, thus Dual-AAE does not need any new network parameters. We also adopt the adversarial mechanism over the latent variables in the O-AE of the proposed framework as in AAE to get better data manifold of the latent variables. In addition, to tackle the mode collapse issue, we introduce the clustering regularization term based on the entropy of the posterior distribution and marginal distribution, which replaces the adversarial training to match the prior distribution of category variables. The basic network architecture of Dual-AAE is shown in Figure~\ref{fig0}. The red arrows represent the data flow direction of D-AE. The observed samples are encoded into the category, style and noise variables by the encoder Q, and then the input samples are reconstructed by the decoder P. D-AE minimizes the reconstruction error of the observed samples to ensure that the hidden variables contain all the information in the data. At the encoding layer of D-AE, the categorical variable is taken into the clustering regularization term to obtain more classification information, while the style variable and random noise are input to the discriminator D to regularize the aggregated Posterior to match its Gaussian prior. The blue arrows represent the data flow direction of O-AE. The category, style and noise variables extracted from the prior distribution are input to the encoder P to generate new samples and then the category and style variables are reconstructed by the decoder Q. O-AE minimizes the reconstruction errors of the category and style variables to reduce information loss in these variables.

Moreover, the accuracy of our proposed Dual-AAE can be further improved by using the \textit{reject option}~\cite{Radu2010Classification}. Since Dual-AAE is a generative clustering model, we can easily obtain the posterior of the category variable. We observe that the maximum clustering probability of many misclassified samples is much less than 1. Therefore, by adding a reject option to the Dual-AAE method, we can select the samples that are difficult to cluster and avoid making decisions on them. These difficult samples can be tackled separately with more effective approaches, such as human supervision\footnote{As is revealed in our experiments, the percentage of such samples is usually small.}.

To evaluate the effectiveness of the proposed Dual-AAE model, we conduct extensive experiments on four benchmark datasets for clustering. The results show that it can achieve superior clustering accuracy over several state-of-the-art algorithms. We also evaluate its clustering performance with a reject option on the MNIST~\cite{lecun1998gradient} dataset, and show that Dual-AAE can almost correctly cluster all samples with only a few samples rejected. Moreover, it is validated in our experiments that Dual-AAE is able to disentangle style and content without supervised information.

Our contributions are summarized as follows.
\begin{itemize}
\item{We propose an unsupervised generative clustering method, Dual-AAE, which extends the AAE by maximizing the mutual information between observed examples and a subset of their latent variables.}
\item{We introduce the clustering regularization term, w.r.t. entropy of $q(y)$ and $q(y|x)$, to encourage the posterior distribution of $\bm{y}$ to match the prior distribution.}
\item{Experimental results show that Dual-AAE outperforms state-of-the-art clustering methods on four benchmarks. Furthermore, by adding a reject option, the clustering accuracy of Dual-AAE can achieve comparable performance of some supervised CNN methods.}
\item{We show the images generated from Dual-AAE conditioned on different values of the latent variables, and illustrate that Dual-AAE can disentangle style and content of images without using any supervised information during training.}
\end{itemize}

\section{Related Work}\label{section2}

A large number of clustering algorithms have been proposed in literature, and generally, they can be classified into two categories: generative clustering methods and discriminative clustering methods. Classical generative clustering methods include $K$-means \cite{macqueen1967some}, GMM \cite{hastie2009overview} and their variants \cite{Xu2017kmeans,Yu2012kmeans,ye2008discriminative,liu2010gaussian,Gebru2016cluster}. They extract classification information by estimating the data distribution. Most of them have to calculate distances in the original data space, thus do not work well in cases of high data dimension, such as image data, due to Curse of Dimensionality \cite{Weruaga2015gmm,Cherian2016Bayesian}. Some clustering methods use kernels to map the original data to a high dimensional feature space without knowing the explicit mapping function, and are hardly applicable to large-scale data since their efficiency heavily depends on the number of samples. Discriminative clustering methods aim to directly identify the categories. Spectral clustering and its variants \cite{ng2002spectral,Liu2017spectral} are among the most popular ones, which use the Laplacian spectra of the similarity matrix to map the original data to a low-dimensional space before clustering. Normalized cuts (N-Cuts)~\cite{Shi2000Ncut} can obtain globally optimized clustering results in the view of graph partition.

In recent years, deep learning has boosted the development of clustering models. Current deep clustering methods are divided to deep generative ones and deep discriminative ones\footnote{See~\cite{Min2018A} for a review of deep discriminative methods.}. Deep generative methods can be further classified into two categories: VAE based models and Generative Adversarial Network (GAN)~\cite{goodfellow2014generative} based models. In the first category, Variational deep embedding (VaDE)~\cite{jiang2017variational} adopts a data generative procedure, which combines VAE and GMM together. Gaussian mixture VAE (GMVAE)~\cite{dilokthanakul2016deep} uses the Gaussian mixture distribution as a prior distribution of the latent variables. In the second category, CatGAN~\cite{Springenberg2016CatGAN} learns a discriminative classifier in the adversarial network and then maximizes the mutual information between observed samples and their predicted category distribution. InfoGAN~\cite{chen2016infogan} introduces the structured latent variable into the random variable of GAN and maximizes mutual information between the structured latent variable and the generated sample. Dual-AAE also has the regularization term to maximize mutual information as in InfoGAN, but there are two key differences in the training process. First, Dual-AAE maximizes mutual information between observed examples and their encodings by an AE while InfoGAN maximizes mutual information between the structured latent variables and the generated samples by a GAN. The clustering performance of InfoGAN depends on the quality of the generated samples, but it is far more difficult to train GAN than AE. Second, Dual-AAE introduces the clustering regularization term to avoid mode collapse, which often occurs in the training of GAN, and to further improve the clustering accuracy. Please see Table \ref{table2} for more details.

\section{Dual Adversarial Auto-Encoders}\label{section3}
In this section we elaborate on our Dual-AAE algorithm. We first introduce the generative process, then give algorithm details, and finally make further analysis on the variational inference in Dual-AAE.

\subsection{Generative Process}

An unsupervised clustering task aims to estimate the posterior of the category variable $p(\bm{y}|\bm{x})$. It is challenging since the classification information is always entangled with other information such as style and background.

Inspired by human cognition mechanism that many complex events are decomposed naturally into a series of isolated and hierarchical concepts \cite{Botvinick2008Hierarchical}, we encode a datum into three parts: the category variable $\bm{y}$, the style variable $\bm{h}$ and the random noise $\bm{z}$. Here the category variable and the style variable are expected to contain the most representative information about the input datum.

In this paper, we consider such a generative model:
\begin{equation}
\label{gen}
p(\bm{x},\bm{y},\bm{h},\bm{z})=p(\bm{x}|\bm{y},\bm{h},\bm{z})p(\bm{y})p(\bm{h})p(\bm{z}).
\end{equation}
Assume both priors of the style variable $\bm{h}$ and the random noise $\bm{z}$ are standard normal distribution and the prior of the category variable $\bm{y}$ is Multinoulli distribution. The observed samples are generated by
\begin{equation}
\begin{split}
\label{eq4}
\bm{y}\sim \mbox{\textit{Mult}}(&\bm{\pi}),\bm{\pi}=(\pi_{1},\pi_{2},\cdots,\pi_{K})\\
\bm{h}&\sim\mathcal{N}(0,\bm{I}_{1})\\
\bm{\bm{z}}&\sim\mathcal{N}(0,\bm{I}_{2})\\
\bm{x}&= G(\bm{y},\bm{h},\bm{z};\bm{\theta}),
\end{split}
\end{equation}
where $K$ represents the number of clusters and is given in advance, and $\pi_{k}$ is the prior of the category variable, $\pi_{k}=1/K$. In particular, the observed sample $\bm{x}$ is generated by a deep neural network $G(\bm{y},\bm{h},\bm{z};\bm{\theta})$.

We expect to obtain the independent posterior distribution, e.g., $p(\bm{y},\bm{h},\bm{z}|\bm{x})=p(\bm{y}|\bm{x})p(\bm{h}|\bm{x})p(\bm{z}|\bm{x})$, by inferring the generative model Eq.~(\ref{gen}). This means the input sample can be disentangled into independent classification information, style information and random noise in order to complete the clustering task.

\subsection{Dual-AAE Algorithm}

As aforementioned, the category variable $\bm{y}$ and the style variable $\bm{h}$ are expected to contain the most representative information about the data. For simplicity, we denote their union as the structured latent variable $\bm{c}$, namely, $\bm{c}=[\bm{y},\bm{h}]$. In information theory, the mutual information $I(X,Y)$ can be used to measure how much information $X$ contains about $Y$. Mutual information can be expressed as
\begin{equation*}
\label{eq5}
I(X,Y)=H(X)-H(X|Y)=H(Y)-H(Y|X),
\end{equation*}
where $H(X)$ denotes the entropy of $X$, and $H(X|Y)$ denotes the conditional entropy of $X$ given $Y$. We then consider extracting the structured latent component $\bm{c}$ from the observed datum $\bm{x}$ via maximizing $I(\bm{x},\bm{c})$.

The objective function of Dual-AAE is formulated to maximize the likelihood function of the observed data with an additional mutual information regularization term:
\begin{equation}
\label{eq6}
\max_{\bm{\theta}} \mathbb{E}_{\bm{x}\sim p_{d}(\bm{x})}[\log p(\bm{x})]+\lambda_{1} I(\bm{x},\bm{c}),
\end{equation}
where $\bm{\theta}$ denotes all the network parameters. However, the mutual information term $I(\bm{x},\bm{c})$ is difficult to  optimize directly since neither the posterior distribution $p(\bm{c}|\bm{x})$ nor the distribution $p(\bm{x}|\bm{c})$ is known. Fortunately, we can use variational inference to obtain a lower bound of $I(\bm{x},\bm{c})$.

For the likelihood function term, let $q(\bm{y},\bm{h},\bm{z}|\bm{x})$ denote the auxiliary distribution to approximate the real posterior $p(\bm{y},\bm{h},\bm{z}|\bm{x})$. The variational process is formulated as
\begin{equation}
\begin{split}
\label{eq7}
&\quad\; \mathbb{E}_{\bm{x}\sim p_{d}(\bm{x})}[\log p(\bm{x})]\\
&=\mathbb{E}_{\bm{x}\sim p_{d}(\bm{x})}[KL(q(\bm{y},\bm{h},\bm{z}|\bm{x})\parallel p(\bm{y},\bm{h},\bm{z}|\bm{x}))\\
&+\mathbb{E}_{q(\bm{y},\bm{h},\bm{z}|\bm{x})}[\log p(\bm{x}|\bm{y},\bm{h},\bm{z})]\\
&-KL(q(\bm{y},\bm{h},\bm{z}|\bm{x})\parallel p(\bm{y},\bm{h},\bm{z}))]\\
&\geq\mathbb{E}_{\bm{x}\sim p_{d}(\bm{x})}[\mathbb{E}_{q(\bm{y},\bm{h},\bm{z}|\bm{x})}[\log p(\bm{x}|\bm{y},\bm{h},\bm{z})]\\
&-KL(q(\bm{y},\bm{h},\bm{z}|\bm{x})\parallel p(\bm{y},\bm{h},\bm{z}))].\\
\end{split}
\end{equation}

For the mutual information term, we derive the variational process as
\begin{equation}
\begin{split}
\label{eq8}
I(\bm{x},\bm{c})&=H(\bm{c})-\mathbb{E}_{p(\bm{x},\bm{c})}[-\log p(\bm{c}|\bm{x})]\\
&=H(\bm{c})+\mathbb{E}_{\bm{x}\sim p(\bm{x})}[KL(p(\bm{c}|\bm{x})\parallel q(\bm{c}|\bm{x}))]\\
&+\mathbb{E}_{\bm{c}\sim p(\bm{c}),\bm{x}\sim p(\bm{x}|\bm{c})}\log q(\bm{c}|\bm{x})\\
&\geq H(\bm{c})+\mathbb{E}_{\bm{c}\sim p(\bm{c}),\bm{z}\sim p(\bm{z}),\bm{x}\sim p(\bm{x}|\bm{c},\bm{z})}\log q(\bm{c}|\bm{x}),\\
\end{split}
\end{equation}
where $p(\bm{c})$ is the prior distribution of the structured latent variable, and $p(\bm{c})=p(\bm{y})p(\bm{h})$. In this paper, we fix the distribution of $\bm{c}$ in Eq.~\eqref{eq8}, so $H(\bm{c})$ is a constant and omitted hereinafter for simplicity. In this variational inference process, the same auxiliary distribution $q(\bm{c}|\bm{x})$ is used to estimate $p(\bm{c}|\bm{x})$, which means no additional variational distribution is required.

Since both $KL(p(\bm{c}|\bm{x})\parallel q(\bm{c}|\bm{x}))$ and $KL(q(\bm{y},\bm{h},\bm{z}|\bm{x})\parallel p(\bm{y},\bm{h},\bm{z}|\bm{x}))$ are non-negative, the evidence lower bound $\mathcal{L}_{D-ELBO}(\bm{x})$ of Dual-AAE can be represented as
\begin{equation}
\begin{split}
\label{eq10}
\mathcal{L}_{D-ELBO} &= \mathbb{E}_{\bm{x}\sim p_{d}(\bm{x})}[\mathbb{E}_{q(\bm{y},\bm{h},\bm{z}|\bm{x})}\log p(\bm{x}|\bm{y},\bm{h},\bm{z})]\\
&+\lambda_{1}\mathbb{E}_{\bm{c}\sim p(\bm{c}),\bm{z}\sim p(\bm{z}),\bm{x}\sim p(\bm{x}|\bm{c},\bm{z})}[\log q(\bm{c}|\bm{x})]\\
&-\mathbb{E}_{\bm{x}\sim p_{d}(\bm{x})}[KL(q(\bm{y},\bm{h},\bm{z}|\bm{x})\parallel p(\bm{y},\bm{h},\bm{z}))],\\
\end{split}
\end{equation}
in which the first two terms are named the reconstruction terms and the last term is the KL divergence term.

As shown in Figure~\ref{fig0}, we use the deep networks Q and P to represent $q(\bm{y},\bm{h},\bm{z}|\bm{x})$ and $p(\bm{x}|\bm{y},\bm{h},\bm{z})$, respectively. For the reconstruction term, the first term can be viewed as the reconstruction loss of the observed datum $\bm{x}$ through O-AE, and the second as the reconstruction loss of the structured latent variable $\bm{c}$ through D-AE. In particular, the encoder and the decoder of D-AE can be represented by the decoder and the encoder of O-AE, respectively. For the KL divergence term, we use the same strategy as AAE to replace it with an adversarial training mechanism, which encourages $q(\bm{y},\bm{h},\bm{z})$ to match the prior $p(\bm{y},\bm{h},\bm{z})$. In this paper, we use the loss of Wasserstein GAN~\cite{arjovsky2017wasserstein}. Therefore, the KL divergence term is replaced by
\begin{equation}
\begin{split}
\label{eq13}
\min_{\bm{\theta}_{Q}}&\max_{\bm{\theta}_{D}}\:\mathbb{E}_{\bm{y},\bm{h},\bm{z}\sim p(\bm{y},\bm{h},\bm{z})}[D(\bm{y},\bm{h},\bm{z})]\\
+\mathbb{E}&_{\bm{x}\sim p_{d}(\bm{x})}[-D(q(\bm{y},\bm{h},\bm{z}|\bm{x}))],
\end{split}
\end{equation}
where $\bm{\theta}_{Q}$ and $\bm{\theta}_{D}$ denote the network parameters of the deep networks Q and D, respectively.

By combining Eq.~\eqref{eq10} and \eqref{eq13}, we expend the evidence lower bound of Dual-AAE as
\begin{equation}
\begin{split}
\label{eq15}
&\min_{\bm{\theta}_{Q},\bm{\theta}_{P}}\max_{\bm{\theta}_{D}}\mathbb{E}_{\bm{x}\sim p_{d}(\bm{x})}\mathbb{E}_{q(\bm{y},\bm{h},\bm{z}|\bm{x})}[-\log p(\bm{x}|\bm{y},\!\bm{h},\!\bm{z})]\\
&\quad\quad\quad\quad\;+\lambda_{1} \mathbb{E}_{\bm{c}\sim p(\bm{c}),\bm{z}\sim p(\bm{z}),\bm{x}\sim p(\bm{x}|\bm{c},\bm{z})}[-\log q(\bm{c}|\bm{x})]\\
&\quad\quad\quad\quad\;+\mathbb{E}_{\bm{x}\sim p_{d}(\bm{x})}[-D(q(\bm{y},\bm{h},\bm{z}|\bm{x}))]]\\
&\quad\quad\quad\quad\;+\mathbb{E}_{\bm{y},\bm{h},\bm{z}\sim p(\bm{y},\bm{h},\bm{z})}[D(\bm{y},\bm{h},\bm{z})],
\end{split}
\end{equation}
where $\bm{\theta}_{P}$ denotes the network parameters of the deep networks P.

\begin{figure}[htb]
\centering{\scalebox{0.35}{\includegraphics{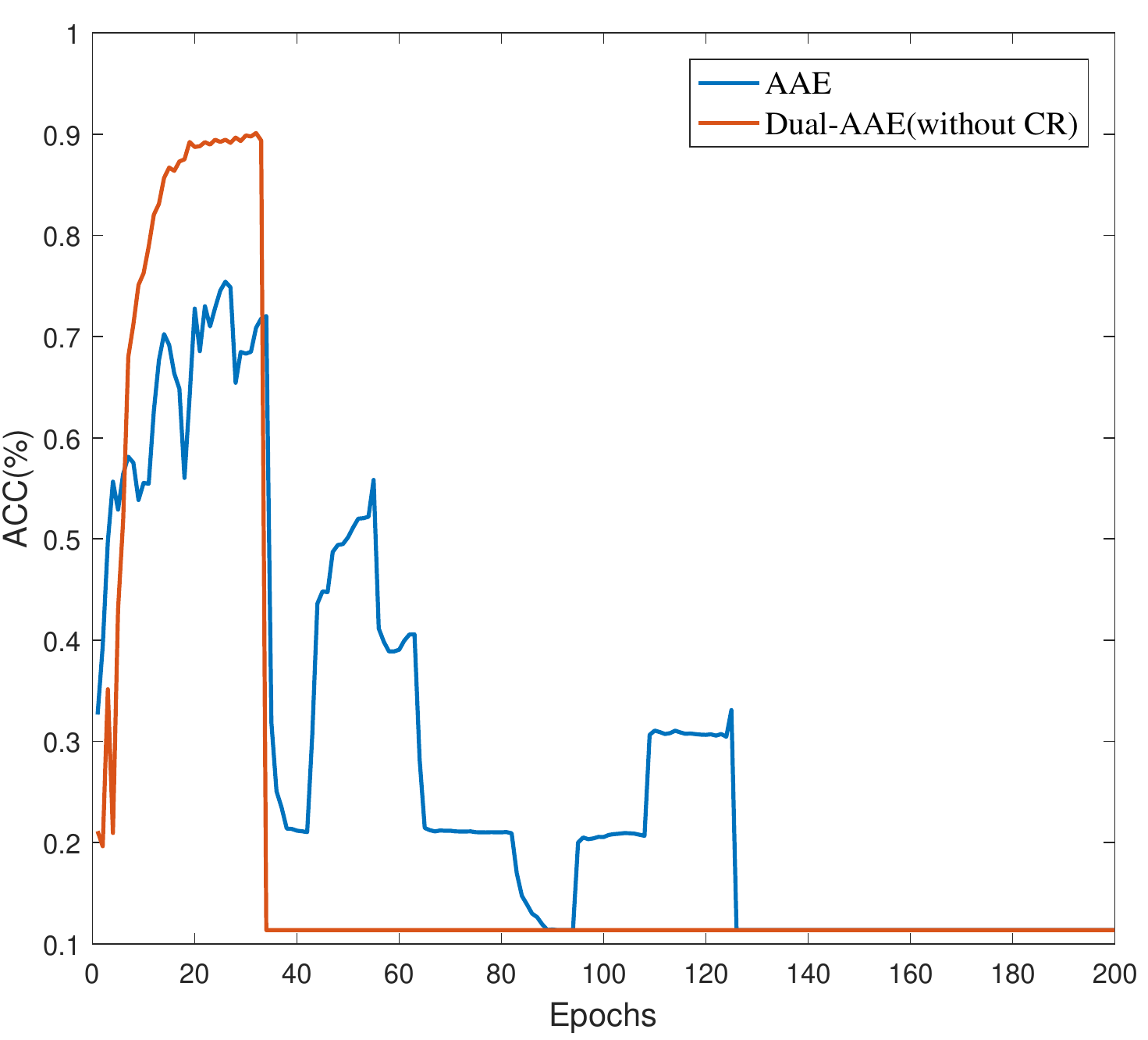}}
\caption{Clustering accuracy w.r.t. number of epochs when AAE and Dual-AAE (without CR) go through mode collapse. Better viewed in color.}
\label{fanli_m}}
\end{figure}

Note that the adversarial training mechanism may suffer mode collapse and get poor clustering results when dealing with the discrete variable $y$. In Figure~\ref{fanli_m}, we show the clustering performance when AAE and Dual-AAE (w.r.t. Eq.~\eqref{eq15}) go through mode collapse. One may find the dataset is divided into one category consequently.

To address the mode collapse issue, we now provide a new clustering regularization (CR) term for Dual-AAE. We introduce two heuristic criteria to encourage $q(\bm{y})$ to match the prior distribution $p(\bm{y})$, instead of using the adversarial training manner in AAE. In Eq. \eqref{eq4}, we define the prior distribution of $\bm{y}$ as a multinoulli distribution with classes balanced. In order to match $p(\bm{y})$, two criteria should be satisfied: 1) when sampling a sample $y_{i}$ from the conditional distribution $q(\bm{y}|\bm{x})$, $y_{i}$ should belong to a certain class; 2) the aggregated posterior $q(\bm{y})$ should be class-balanced. To satisfy the first criterion, the entropy of $q(\bm{y}|\bm{x})$ should be low, while to satisfy the second, the entropy of $q(\bm{y})$ should be high. Accordingly, we can formulate these two criteria as
\begin{equation}
\begin{split}
\label{eq14}
\min_{\bm{\theta}_{Q}}-H[q(\bm{y})]+\lambda_{2}\mathbb{E}_{\bm{x}\sim p_{d}(\bm{x})}[H[q(\bm{y}|\bm{x})]],
\end{split}
\end{equation}
where the hyper-parameter $\lambda_{2}$ is used for balancing these two items. Eq. \eqref{eq14} encourages the category variable $\bm{y}$ to capture the classification information from the data. Note that though CatGAN~\cite{Springenberg2016CatGAN} uses a similar formula, there exists a significant difference between CatGAN and Dual-AAE: CatGAN uses the formula to guide the training of GAN while Dual-AAE uses Eq. \eqref{eq14} to encourage $q(\bm{y})$ to match the prior distribution $p(\bm{y})$.

We use Eq. \eqref{eq14} to replace the adversarial training of $y$ in Eq. \eqref{eq15}, and the objective function of Dual-AAE is finally formulated as
\begin{equation}
\begin{split}
\label{eqdaae}
&\min_{\bm{\theta}_{Q},\bm{\theta}_{P}}\max_{\bm{\theta}_{D}}\mathbb{E}_{\bm{x}\sim p_{d}(\bm{x})}\mathbb{E}_{q(\bm{y},\bm{h},\bm{z}|\bm{x})}[-\log p(\bm{x}|\bm{y},\!\bm{h},\!\bm{z})]\\
&+\lambda_{1} \mathbb{E}_{\bm{c}\sim p(\bm{c}),\bm{z}\sim p(\bm{z}),\bm{x}\sim p(\bm{x}|\bm{c},\bm{z})}[-\log q(\bm{c}|\bm{x})]\\
&+\mathbb{E}_{\bm{x}\sim p_{d}(\bm{x})}[-D(q(\bm{h},\bm{z}|\bm{x}))]]+\mathbb{E}_{\bm{h},\bm{z}\sim p(\bm{h},\bm{z})}[D(\bm{h},\bm{z})]\\
&-H[q(\bm{y})]+\lambda_{2}\mathbb{E}_{\bm{x}\sim p_{d}(\bm{x})}[H[q(\bm{y}|\bm{x})]].
\end{split}
\end{equation}

We call the first two terms as reconstruction loss, the next two terms as adversarial loss, and the last two terms as CR term. The reconstruction loss of O-AE encourages Dual-AAE to explain the input data well while the reconstruction loss of D-AE encourages the structure latent variable $\bm{c}$ to extract as much information as possible from the data. The adversarial loss uses an adversarial training procedure to regularize the aggregated posterior of style latent variable $\bm{h}$ and random noise $\bm{z}$ to match the \textit{Gaussian} prior. The CR term minimizes the entropy of $q(\bm{y}|\bm{x})$ while maximizing the entropy of $q(\bm{y})$, thus it encourages the category variable $\bm{y}$ to capture the classification information from the input data. All these loss functions can be calculated efficiently, and the Dual-AAE network parameters can be optimized by the back propagation algorithm.

After the optimization process of Dual-AAE is completed, we can obtain the variational posterior distribution $q(\bm{y}|\bm{x}$). Dual-AAE assigns the input data to the cluster with the highest posterior probability as the clustering result. However, for some easily confused samples, even the best supervised classification algorithms cannot address them effectively. We therefore intuitively propose to pick them out by adding a reject option. With the posterior of the category variable estimated by Dual-AAE, we can set a threshold $\gamma$ and easily reject the input sample $\bm{x}$ which has a largest posterior distribution less than or equal to $\gamma$. Then, the rejected samples will be processed separately with more effective and supervised approaches, such as manual annotation.

\subsection{Further Analysis of the Dual-AAE Variational Inference}

Like VAE, the proposed Dual-AAE also uses the variational inference to learn a generative model. Compared with the standard VAE, there are two major modifications in variational inference of Dual-AAE.

First, unlike VAE that only maximizes the likelihood function, our Dual-AAE simultaneously maximizes the likelihood function and mutual information between $\bm{x}$ and $c$. Eq.~\eqref{eq8} shows the variational process of the mutual information term. When the evidence lower bound of $I(\bm{x},\bm{c})$ reaches its maximum value $H(\bm{c})$, the KL divergence between the true distribution $p(\bm{c}|\bm{x})$ and the variational distribution $q(\bm{c}|\bm{x})$ becomes zero, which means Dual-AAE encourages the variational distribution of the structural latent variable to approach the true posterior.

Second, Dual-AAE has a clustering regularization term. Eq.~\eqref{eq7} shows the variational process of applying the standard VAE to clustering tasks. When we maximize the variational lower bound, the anti-clustering term in the evidence lower bound of VAE, i.e., $-KL(q(\bm{y}|\bm{x})\parallel p(\bm{y}))$, will reduce the KL divergence between the posterior distribution of $\bm{y}$ and the uniform prior. Thus VAE needs to be modified to accommodate the clustering tasks. AAE uses an adversarial training mechanism to encourage the aggregated posterior of the latent variables to match the prior distribution. However, we find when dealing with the discrete variable $\bm{y}$, the adversarial training mechanism may fall into mode collapse and get poor clustering results. Comparatively, Dual-AAE introduces the clustering regularization term to encourage the posterior of $\bm{y}$ to match the prior distribution, which enables the model to escape from the threat of model collapse.

\section{Experiment}\label{section4}

In this section, we evaluate the clustering performance of Dual-AAE from several views. We first test its accuracy and efficiency comparing with other methods, and show its performance can be further lifted by adding the reject option. Then we evaluate its performance in case of unknown cluster number. After that, we demonstrate the generated samples of Dual-AAE. Finally, we verify that Dual-AAE can learn disentangled and meaningful representations.

\subsection{Experimental Setups}

\begin{table}[tb]
\caption{Clustering accuracy ($\%$) on all datasets.}
\centering
\renewcommand{\tabcolsep}{0.5pc} % enlarge column spacing
\renewcommand{\arraystretch}{1.2} % enlarge line spacing
\label{table2}
\begin{threeparttable}
\begin{tabular}{c|c|c|c|c}
\hline
Method & MNIST & HHAR & STL-10 & REUTERS\\
\hline
K-means & 53.55 & 60.08 & 64.86& 54.04 \\
\hline
N-Cuts   & 44.83  &  53.60  &  64.23 & N/A\\
\hline
GMM & 53.73 & 60.34 & 72.44 & 55.81\\
\hline
VAE+GMM & 72.94 & 68.02 & 78.86 & 70.98 \\
\hline
DEC & 84.30 & 79.86 & 80.62 & 75.63\\
\hline
GMVAE  & 88.54 & - & - & -\\
\hline
DEPICT  & 96.50 & - & - & -\\
\hline
AAE & 84.10 & 83.77 & 81.24 & 75.12\\
\hline
CatGAN & 84.61 & 80.77 & 83.08 & 75.9\\
\hline
InfoGAN & 96.33 & 83.37 & 82.14 & 75.33\\
\hline
VaDE & 94.46 & 84.45 & 84.46 & 79.38\\
\hline
Dual-AAE(without CR) & 97.04 & 85.94 & 86.85 & 79.56\\
\hline
{\bfseries Dual-AAE} & {\bfseries 97.86} & {\bfseries 86.79} & {\bfseries89.15} & {\bfseries81.45}\\
\hline
\end{tabular}
\end{threeparttable}
\end{table}

All experiments are performed on four datasets, MNIST~\cite{lecun1998gradient}, HHAR~\cite{stisen2015smart}, STL-10~\cite{coates2011analysis} and REUTERS~\cite{Lewis2004RCV1}, which are widely used for testing clustering algorithms. The MNIST dataset includes 70,000 handwritten digital images, each with a size of $28\times28$. No pre-processing is applied to the data except for scaling to the numerical range $[0,1]$. The HHAR dataset consists of 10,299 sensor records about 6 different human activities, and the dimension of each sample is 561. The STL-10 dataset is composed of 13,000 color images over ten categories, and each image is of size 96$\times$96. Following the settings in VaDE~\cite{jiang2017variational}, we extract image features by ResNet-50 \cite{he2016deep} as the input data, and the dimension of Res-features is 2048. The REUTERS dataset consists of 810,000 English news stories. Following DEC~\cite{xie2016unsupervised} and VaDE~\cite{jiang2017variational}, we use 685,071 samples from 4 common categories, and each story is represented by a 2000-dimensional tf-idf feature.

In our Dual-AAE, we use deep neural networks to instantiate the encoder, decoder, and discriminator. In the objective function of Dual-AAE, we introduce two hyper-parameters, $\lambda_1$ and $\lambda_2$. Since the only information we know in unsupervised clustering is the observed datum $\bm{x}$, the reconstruction of $\bm{x}$ should be our primary objective, and the reconstruction of $\bm{c}$ can be viewed as a regularization term, thus $\lambda_1$ should be set to a small number. In our experiments, we set $\lambda_1=0.1$. In the clustering loss, we observe that cluster balance has a greater impact on improving the clustering performance, and we empirically set $\lambda_2=0.5$. For more parameter settings used in our experiments, please see Appendix for details.

We use a standard evaluation metric, clustering accuracy (ACC) \cite{xu2003document}, to evaluate clustering performance, defined as
\begin{equation*}
ACC = \sum_{i=1}^{N}\delta\left(y^{(i)},\mbox{\textit{map}}(l^{(i)})\right)/N,
\end{equation*}
where $N$ is the number of samples, $\bm{y}$ is the ground truth label and $\bm{l}$ is the resolved cluster label. The function $\mbox{\textit{map}}(l^{(i)})$ is a one-to-one mapping, used for mapping each cluster label $l^{(i)}$ to the ground truth label. The best mapping can be found by the Kuhn-Munkres (KM) algorithm \cite{lovasz2009matching}. $\delta(x,y)$ is the delta function, which equals 1 only if $x=y$ and 0 otherwise.

\subsection{Performance Comparison with State-of-the-arts}

\begin{figure}[t]
\centering{\scalebox{0.35}{\includegraphics{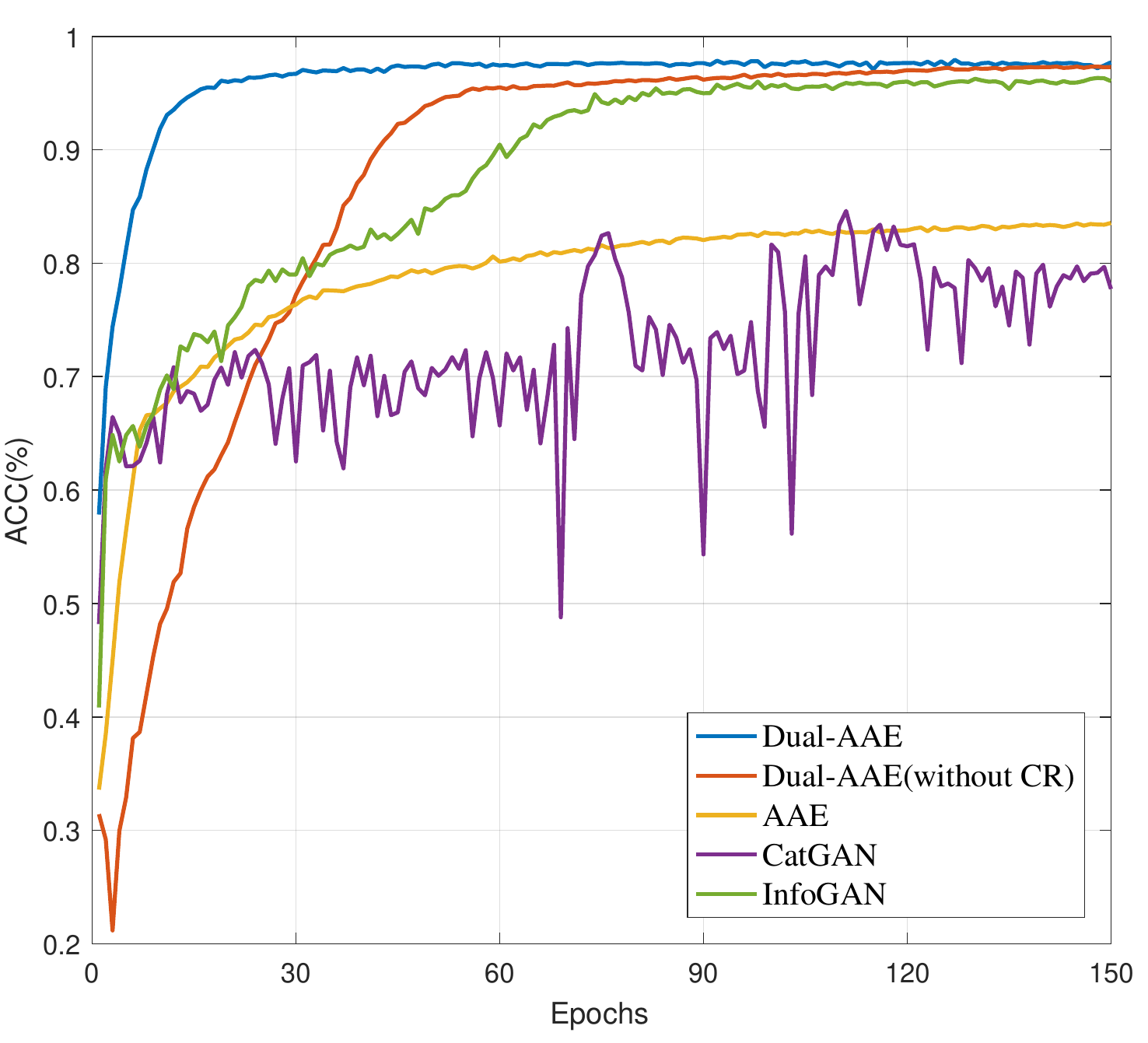}}
\caption{Clustering accuracies on MNIST. Better viewed in color.}
\label{fig:2}}
\end{figure}

\begin{figure*}[htb]
\subfigure[Raw Data]{\label{Fig.sub.em1}
\begin{minipage}[c]{0.3\textwidth} % 0.23  0.5
\centering \scalebox{0.5}{ % 0.42
\includegraphics{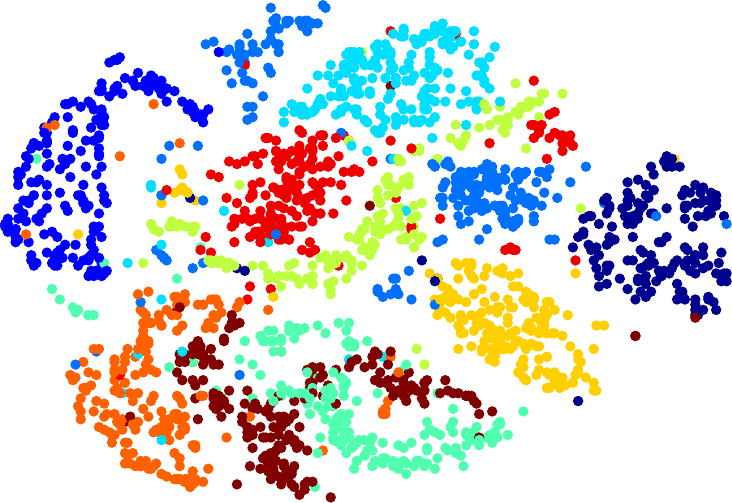}}
\end{minipage}}
\subfigure[AAE]{\label{Fig.sub.em2}
\begin{minipage}[c]{0.3\textwidth}
\centering \scalebox{0.45}{ %0.42
\includegraphics{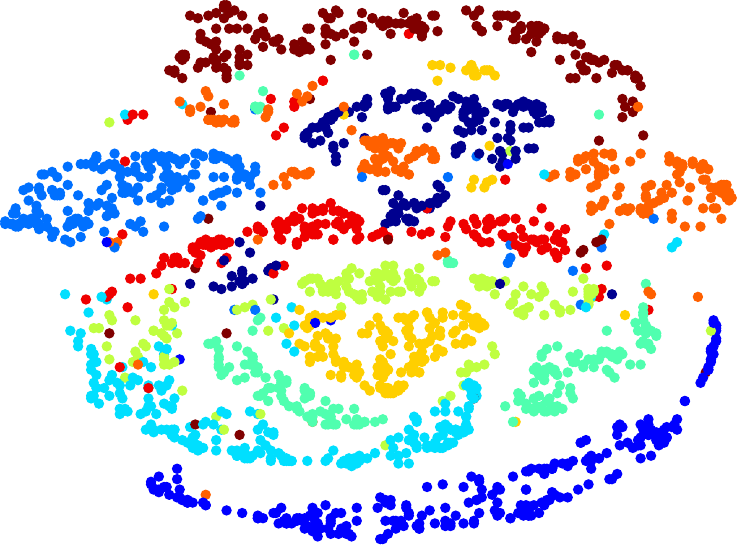}}
\end{minipage}}
\subfigure[Dual-AAE]{\label{Fig.sub.em3}
\begin{minipage}[c]{0.3\textwidth}
\centering \scalebox{0.45}{ %0.42
\includegraphics{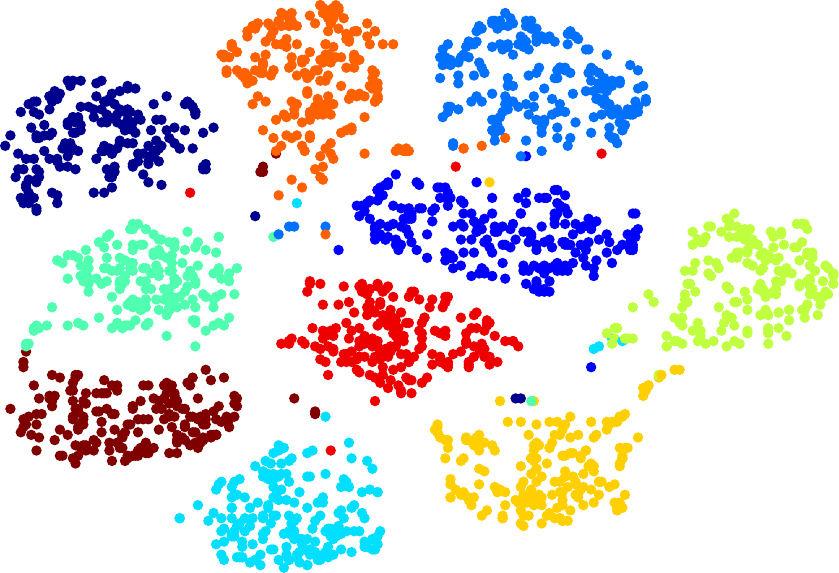}}
\end{minipage}}
\caption{Visualization of the embedding subspaces using the MNIST-test data. (a) Raw data. (b) The embedding subspace of AAE. (c) The embedding subspace of our Dual-AAE. Better viewed in color.}\label{fig:em}
\end{figure*}

We first compare the performance of our proposed Dual-AAE with other state-of-the-art clustering algorithms, including CatGAN~\cite{Springenberg2016CatGAN}, GMVAE~\cite{dilokthanakul2016deep}, DEPICT~\cite{dizaji2017deep}, AAE~ \cite{makhzani2015adversarial}, InfoGAN~\cite{chen2016infogan}, DEC~\cite{xie2016unsupervised} and VaDE~\cite{jiang2017variational}.

We calculate the clustering ACC on all four datasets. For fair comparison, the same network architecture is used to evaluate performance of CatGAN~\cite{Springenberg2016CatGAN}, AAE \cite{makhzani2015adversarial}, InfoGAN \cite{chen2016infogan}, Dual-AAE (without CR) and Dual-AAE. For the rest methods, we cite the best results reported either from the original work or from \cite{jiang2017variational}. Table~\ref{table2} summarizes the quantitative results. On all datasets, Dual-AAE and Dual-AAE (without CR) achieve the best and second best ACC, respectively. These show the benefits of the D-AE and CR terms proposed in this paper. In particular, comparing with the baseline model AAE, the ACC of Dual-AAE increases by 13.76\%, 3.02\%, 7.91\% and 6.33\%, respectively, which are significant improvements of clustering performance.

In addition, we show the clustering accuracy of different models over the number of epochs in Figure~\ref{fig:2}. One can see that the ACC curve of AAE based models is more stable than that of GAN based models. In particular, the performance of Dual-AAE is enhanced rapidly in the first 10 epochs and then it converges to a solution after 20 epochs, while Dual-AAE (without CR) converges to a solution after 60 epochs, which means CR can also make Dual-AAE more efficient.

In Figure \ref{fig:em}, we visualize the embedding subspace of different methods on 2,000 randomly sampled digits from the MNIST-test data. Figure \ref{Fig.sub.em1} visualizes the raw data representations; Figure \ref{Fig.sub.em2} shows the data points in the embedding subspace of AAE; Figure \ref{Fig.sub.em3} shows the feature representations of our Dual-AAE method. For the AAE and Dual-AAE, we extract the feature representations in the penultimate layer rather than the last layer of the encoder. Then, we use t-SNE \cite{Maaten2008Visualizing} to reduce the dimensionality of these feature representations to 2. As shown in Figure \ref{fig:em}, AAE cannot effectively extract the classification information when it is used for clustering tasks. Comparatively, our Dual-AAE provides a more separable embedding subspace and is able to extract better classification information than AAE.

We have shown the clustering performance, when AAE and Dual-AAE(without CR) go through mode collapse, in Figure~\ref{fanli_m}. To further test the role of the CR term in avoiding mode collapse, we follow the experimental scheme in unrolled GAN~\cite{Metz2016UnrollGAN} to count the number of modes covered and calculate the KL divergence between $q(y)$ and $p(y)$. We evaluate AAE, Dual-AAE (without CR) and Dual-AAE on MNIST for demonstration. In each experiment, we repeat the algorithm 10 times and show the mean and standard deviation in Table~\ref{tableCR}. AAE and Dual-AAE(without CR) cannot generate all modes and have large standard deviations, which indicate that they fall into mode collapse at least once. Dual-AAE generates all modes, and both mean and standard deviation values of KL divergence are close to 0. It indicates that the CR term can avoid mode collapse effectively and match the prior distribution well.

\begin{table}[htb]
\caption{The number of modes covered and $KL(q(y)\parallel p(y))$ of AAE, Dual-AAE(without CR) and Dual-AAE on MNIST.}
\centering
\renewcommand{\tabcolsep}{0.33pc} % enlarge column spacing
\renewcommand{\arraystretch}{1.2} % enlarge line spacing
\label{tableCR}
\begin{tabular}{c|c|c|c}
\hline
Method & AAE & Dual-AAE(without CR) & Dual-AAE\\
\hline
Modes generated & 8.10$\pm$2.99 & 8.20$\pm$3.79 & 10$\pm$0 \\
\hline
$KL(q(y)\parallel p(y))$ & 0.38$\pm$0.68 & 0.48$\pm$0.96 & 0.0026$\pm$0.00028 \\
\hline
\end{tabular}
\end{table}

We also test the time complexity of Dual-AAE and compare it with AAE and InfoGAN, the two most related methods. Compared with AAE, Dual-AAE includes a pair of Auto-encoders, O-AE and D-AE, which share the same network parameters. This means Dual-AAE has computation complexity approximately twice that of AAE. We count the training time for an epoch on MNIST, HHAR and STL-10 datasets and all experiments run on a single TITAN Xp GPU card. The results are shown in Table \ref{table4}. Taking the MNIST dataset as an example, the running time for an epoch of Dual-AAE is 1.8 times that of AAE and 1.6 times that of InfoGAN. Considering Dual-AAE converges faster (see more details in Figure \ref{fig:2}), our Dual-AAE achieves a running time complexity comparable to these two baseline methods.

\begin{table}[htb]
\caption{Training time of AAE, InfoGAN and Dual-AAE.}
\centering
\renewcommand{\tabcolsep}{1.4pc} % enlarge column spacing
\renewcommand{\arraystretch}{1.2} % enlarge line spacing
\label{table4}
\begin{tabular}{c|c|c|c}
\hline
Method & MNIST & HHAR & STL-10\\
\hline
AAE & 40.55 & 4.82 & 5.35 \\
\hline
InfoGAN & 46.23 & 3.59 & 3.87 \\
\hline
Dual-AAE & 75.24 & 6.31 & 7.02\\
\hline
\end{tabular}
\end{table}

\subsection{Performance Analysis with Reject Option}

In this group of experiments, we show that the performance of our method can be further lifted by adding a reject option.

\begin{figure}[htb]
\centering{\scalebox{0.3}{\includegraphics{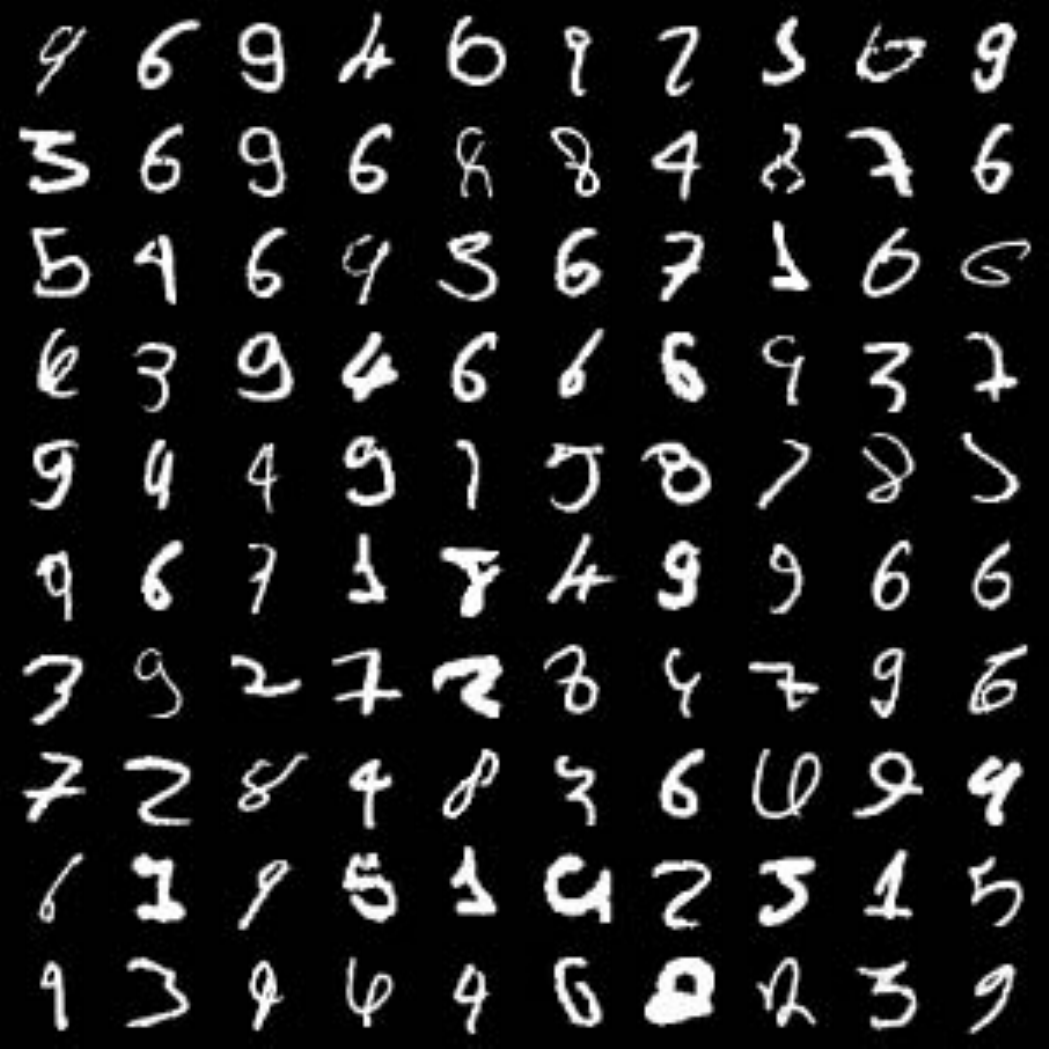}}\caption{Misclassified samples in the MNIST dataset.}\label{fig:3}} % 0.4  0.75
\end{figure}

As can be seen, the clustering accuracy of Dual-AAE on the MNIST dataset reaches 97.86\%, which is competitive to some supervised classification methods. It is observed that the mistakes mainly come from the shape features of the digits. Examples are given in Figure \ref{fig:3}. The 100 misclassified samples shown in the figure can be classified into two categories. One class contains outliers, such as the first sample in line four or the last sample in line five, which are difficult to distinguish even for human beings. The other class can be viewed as a combination of two different digits, e.g. $7$ with a bar nearly in the bottom that can be seen as a combination of $7$ and $2$. We find that the samples of the second class are always equally divided into two clusters, which inspires us to add a reject option to the Dual-AAE algorithm to reduce such error.

\begin{table}[htb]
\caption{Clustering accuracy ($\%$) and rejection rate ($\%$) of Dual-AAE with reject option on MNIST.}
\centering
\renewcommand{\tabcolsep}{1.6pc} % enlarge column spacing
\renewcommand{\arraystretch}{1.2} % enlarge line spacing
\label{table3}
\begin{tabular}{c|c|c}
\hline
Threshold & Rejection rate & ACC\\
\hline
- & 0.00 & 97.86 \\
\hline
0.50 & 0.89 & 98.22 \\
\hline
0.60 & 1.60 & 98.51\\
\hline
0.90 & 8.16 & 99.39\\
\hline
\end{tabular}
\end{table}

We evaluate the clustering performance of the Dual-AAE with reject option on the MNIST dataset. Table \ref{table3} reports the clustering ACC ($\%$) and rejection rate ($\%$) at different thresholds $\theta$. As is shown, with a small rejection threshold, the clustering ACC is significantly improved and only a few samples are rejected. In particular, when the rejection threshold is set as 0.5, the clustering accuracy is already close to the classification accuracy of some supervised classification methods with only 0.89$\%$ of the samples rejected. This indicates that Dual-AAE can effectively improve the clustering ACC with only a small number of samples being rejected.

\subsection{Performance Analysis w.r.t. Unknown Cluster Number}

\begin{figure}[tb]
\subfigure[7 clusters]{\label{Fig.sub.m1}
\begin{minipage}[b]{0.23\textwidth} % 0.23  0.5
\centering \scalebox{0.3}{ % 0.42
\includegraphics{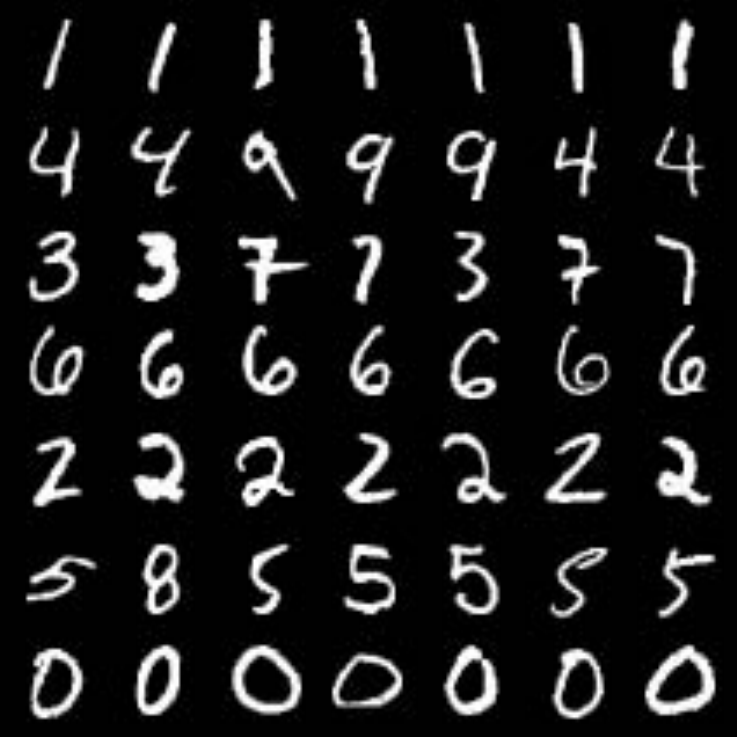}}
\end{minipage}}
\subfigure[12 clusters]{\label{Fig.sub.m2}
\begin{minipage}[b]{0.23\textwidth}
\centering \scalebox{0.47}{ %0.42
\includegraphics{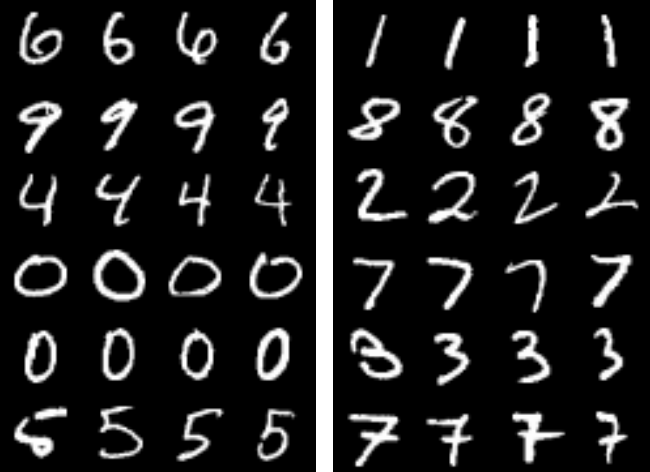}}
\end{minipage}}\\
\caption{Clustering results by Dual-AAE on MNIST while the numbers of clusters are set to 7 and 12. Samples in the same row belong to one cluster.}\label{fig:m712}
\end{figure}

In the above experiments, the number of classes for each dataset is given to Dual-AAE as a prior knowledge, but in some cases it is unknown. We here evaluate the clustering performance of Dual-AAE when the number of clusters is incorrectly specified. Figure \ref{fig:m712} shows the clustering results by Dual-AAE on MNIST dataset when the number of clusters is set to 7 and 12. In Figure \ref{Fig.sub.m1}, we can see when the number of clusters is smaller than the true number of classes, some similar categories will be clustered into one cluster, such as $4$ and $9$, $3$ and $7$. In Figure \ref{Fig.sub.m2}, when the number of clusters is larger than the true number of classes, some digits will be clustered into multi-clusters due to intra-class diversity, such as $7$ and $7$ with a bar in the middle, the fatter $0$ and thinner $0$. It can be seen that when the number of categories is incorrectly set, Dual-AAE can still effectively extract the classification information.

\subsection{Sample Generation Analysis}

In this subsection, we evaluate the ability of Dual-AAE to generate new samples. As a generative clustering model, Dual-AAE has a natural advantage over other generative models (like VAE, GAN) in generating high-quality and active samples from specified clusters.

\begin{figure}[tb]
\subfigure[VAE]{\label{Fig.sub.1-1}
\begin{minipage}[b]{0.23\textwidth} % 0.23  0.5
\centering \scalebox{0.23}{ % 0.42
\includegraphics{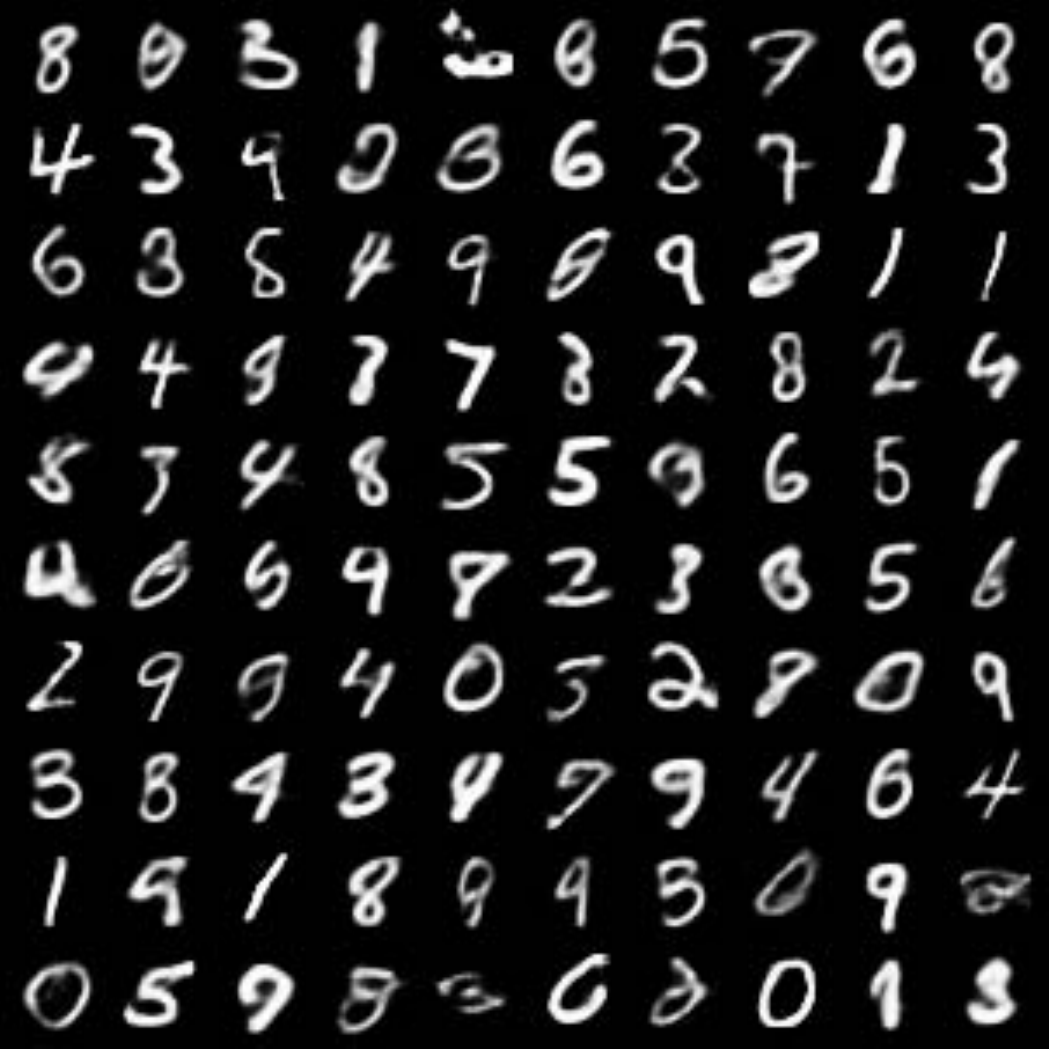}}
\end{minipage}}
\subfigure[GAN]{\label{Fig.sub.1-2}
\begin{minipage}[b]{0.23\textwidth}
\centering \scalebox{0.23}{ %0.42
\includegraphics{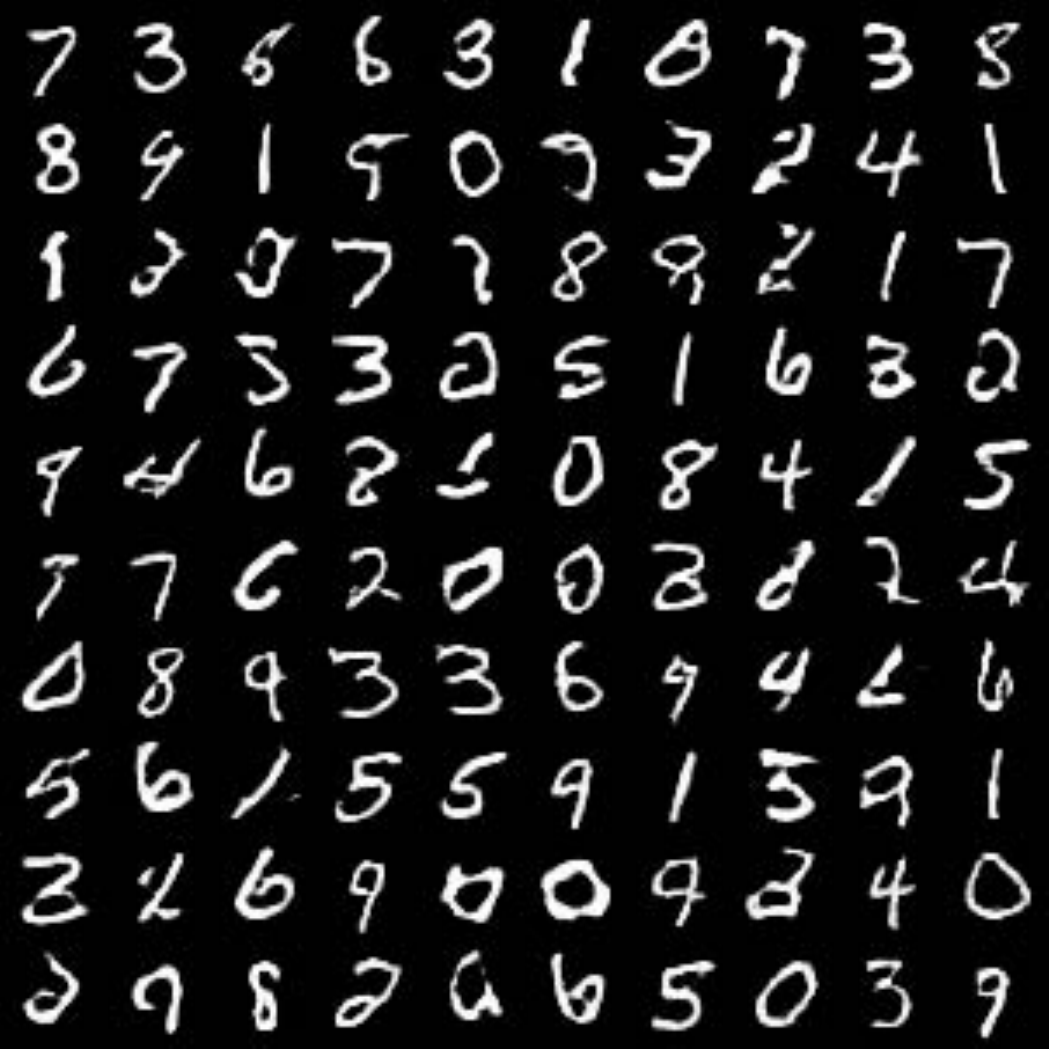}}
\end{minipage}}\\
\subfigure[InfoGAN]{\label{Fig.sub.2-1}
\begin{minipage}[b]{0.23\textwidth}
\centering \scalebox{0.23}{ %0.42
\includegraphics{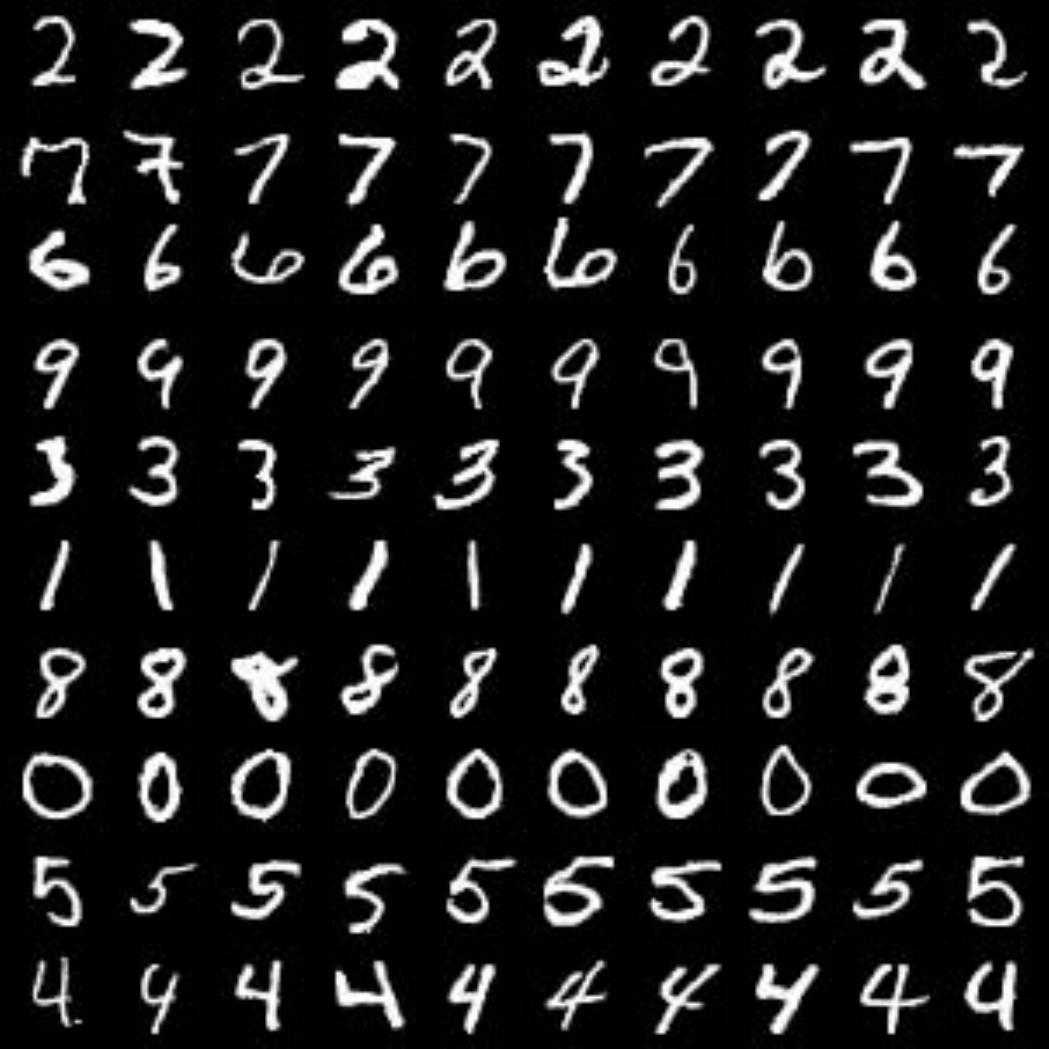}}
\end{minipage}}
\subfigure[Dual-AAE]{\label{Fig.sub.2-2}
\begin{minipage}[b]{0.23\textwidth}
\centering \scalebox{0.23}{ %0.42
\includegraphics{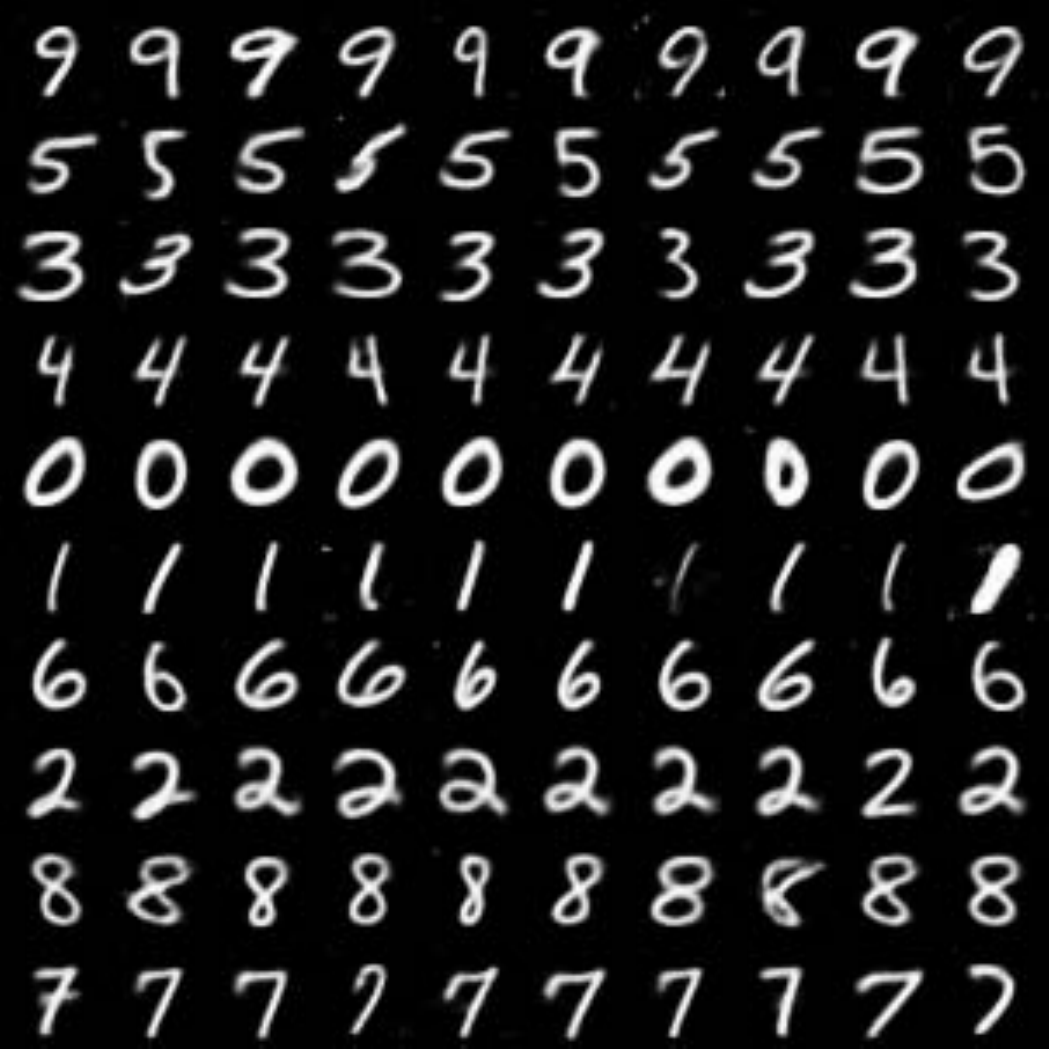}}
\end{minipage}}
\caption{Samples generated from VAE, AAE, InfoGAN and Dual-AAE. Top: We randomly sample from the latent variable space, and all the images are generated by decoding these latent variables by the decoder. Bottom: Images in the same row come from the same cluster.}\label{fig:4}
\end{figure}

In Figure \ref{fig:4}, we show the generated samples by four different generative models: VAE~\cite{kingma2013auto}, GAN~\cite{goodfellow2014generative}, InfoGAN~\cite{chen2016infogan} and Dual-AAE. In Figure \ref{Fig.sub.1-1} and Figure \ref{Fig.sub.1-2}, VAE and GAN cannot generate samples from specified clusters. InfoGAN is able to learn excellent disentangled representations via combining GAN and information theory, and thus can generate samples from specified clusters. We compare the samples generated by InfoGAN and our Dual-AAE. In Figure \ref{Fig.sub.2-1} and \ref{Fig.sub.2-2}, each row contains 10 randomly generated samples from specified clusters. It can be seen clearly that the samples in the same row belong to the same digit, which means our results are comparable to the state-of-the-art generative model InfoGAN. We also see Dual-AAE can generate smooth and diverse digits, which make these images more realistic.

\subsection{Disentangled Representation Analysis}

\begin{figure}[tb]
\subfigure[Varying $h_{1}$ (Width)]{\label{Fig.sub1.1-1}
\begin{minipage}[b]{0.22\textwidth} % 0.23  0.5
\centering \scalebox{0.22}{ % 0.42
\includegraphics{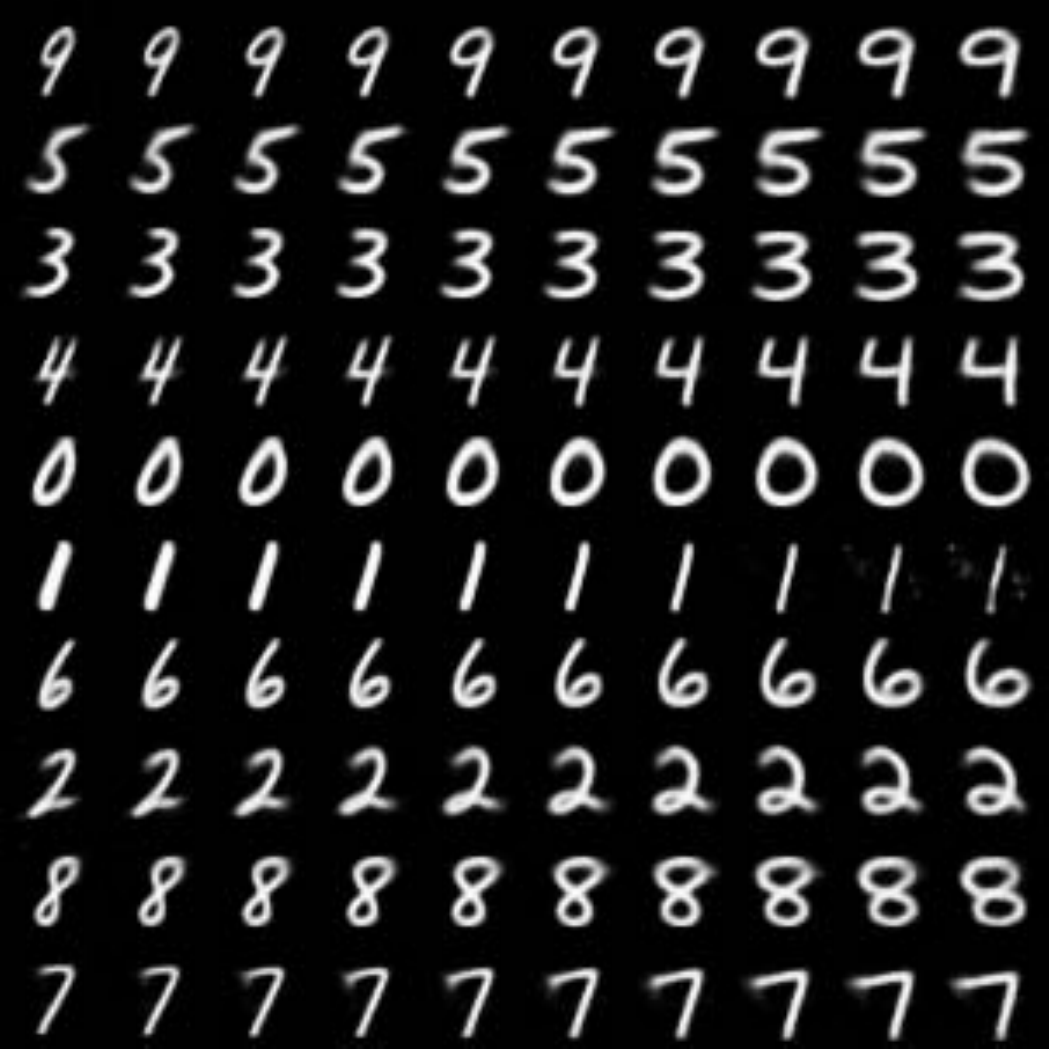}}
\end{minipage}}
\subfigure[Varying $h_{2}$ (Rotation)]{\label{Fig.sub1.1-2}
\begin{minipage}[b]{0.22\textwidth}
\centering \scalebox{0.22}{ %0.42
\includegraphics{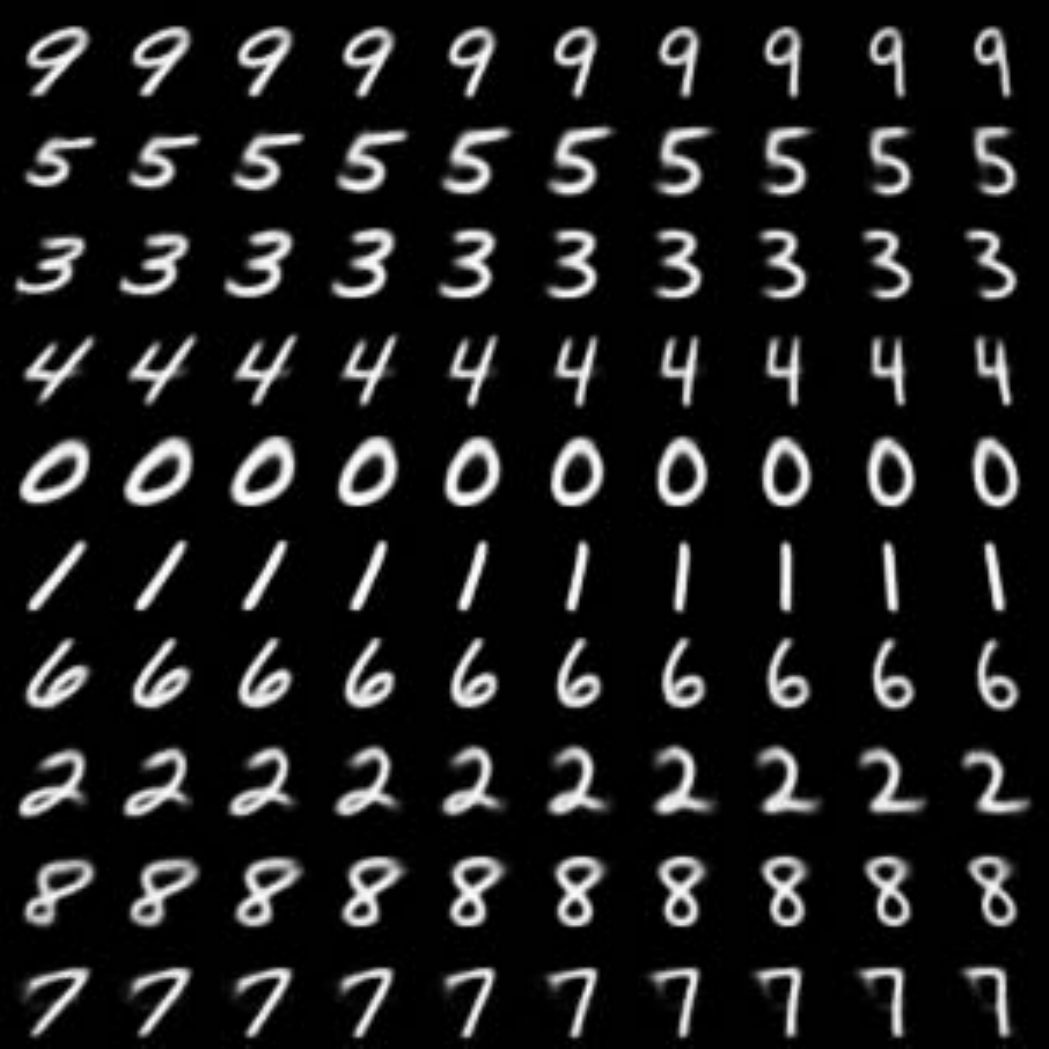}}
\end{minipage}}\\
\subfigure[Varying $h_{3}$ (Centre of gravity)]{\label{Fig.sub1.2-1}
\begin{minipage}[b]{0.22\textwidth}
\centering \scalebox{0.22}{ %0.42
\includegraphics{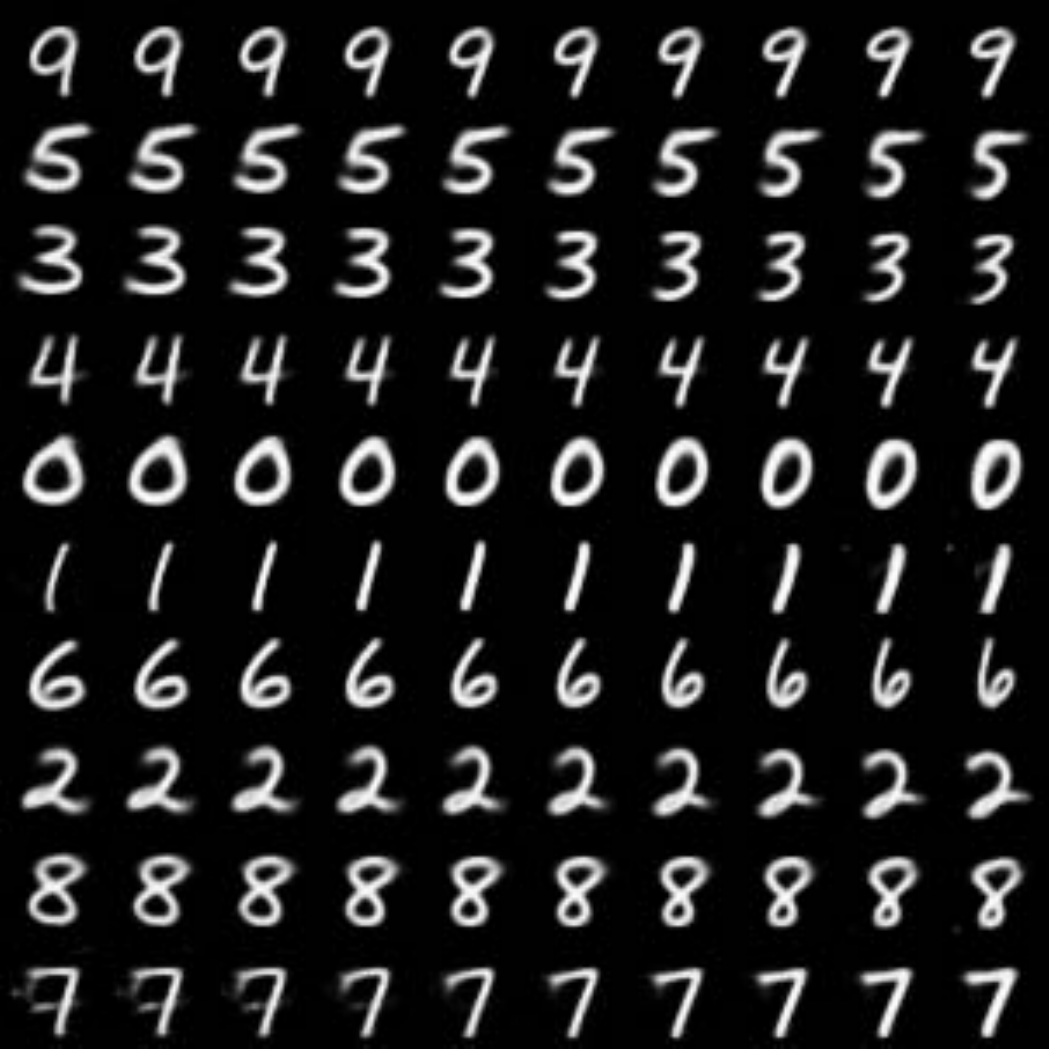}}
\end{minipage}}
\subfigure[Varying $z$]{\label{Fig.sub1.2-2}
\begin{minipage}[b]{0.22\textwidth}
\centering \scalebox{0.22}{ %0.42
\includegraphics{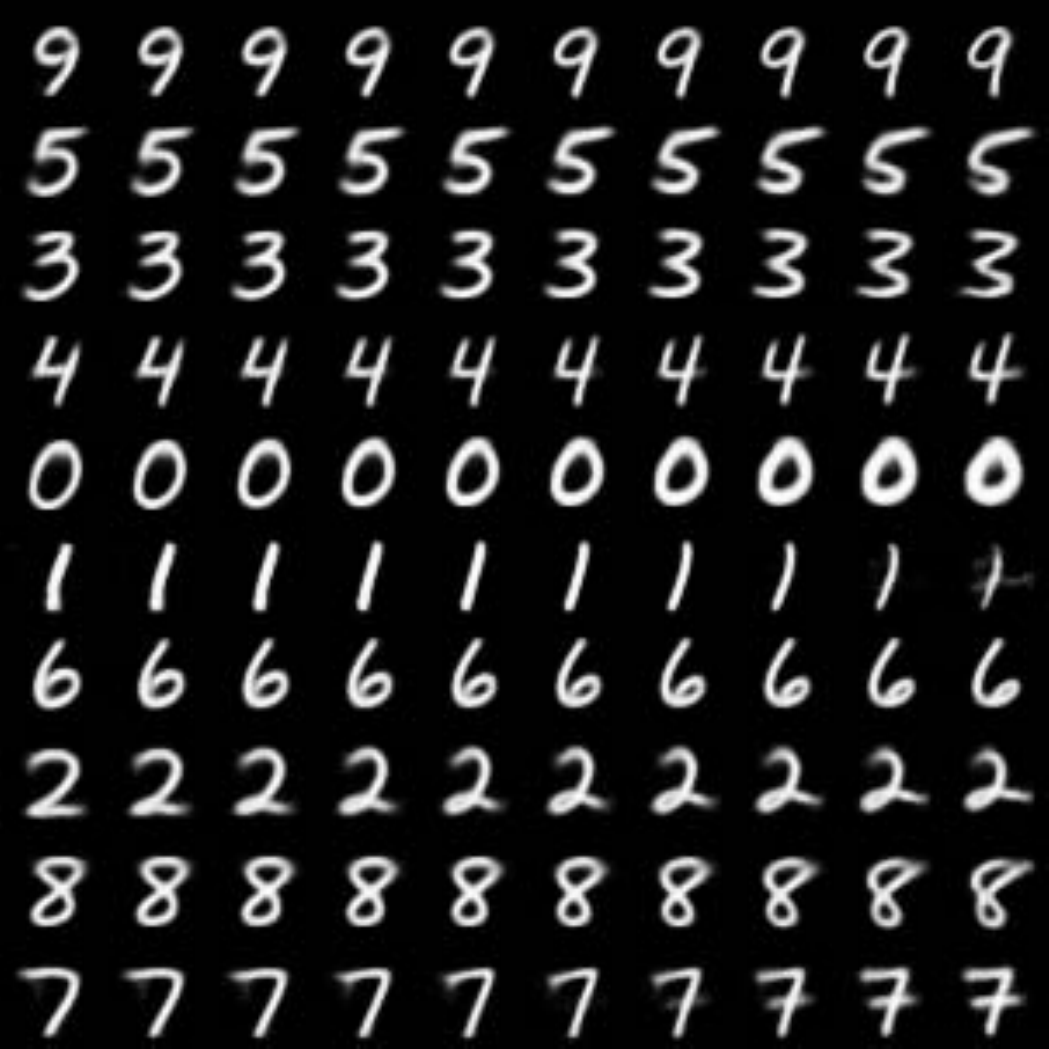}}
\end{minipage}}
\caption{Varying style latent codes and random noise on MNIST. Each figure increases the value of one style variable or the random noise from left to right while fixing other latent variables. In the same column, we vary the category variable and fix the other variables.}\label{fig:5}
\end{figure}

In this subsection, the visualized results on MNIST are provided to show that Dual-AAE can learn interpretable and meaningful representations. In MNIST, samples are encoded as the category variable $\bm{y}$, style variable $\bm{h}$ and random noise $\bm{z}$. Specifically, as shown in Eq. \eqref{eq4}, we assume that the category variable $\bm{y}\sim \mbox{\textit{Mult}}(K=10)$, the style variable $\bm{h}\sim \mathcal{N}(\textbf{0}_1, \bm{I}_1)$, and the random noise $\bm{z}\sim \mathcal{N}(\textbf{0}_2, \bm{I}_2)$, in which the subscripts are used to denote different sizes.

In Figure \ref{fig:5}, we show the category variable $\bm{y}$ can capture the classification information from the MNIST dataset, and changing $\bm{y}$ could always generate different digits. Dual-AAE can extract the classification information so well that we achieve the-state-of-art clustering accuracy.

The style variable $\bm{h}$ can capture the style information in the MNIST dataset. In Dual-AEE, three style variables control three different styles: width, rotation, and center of gravity, respectively. In order to show the extracted style information more clearly, in each subplot of Figure \ref{fig:5}, we take the values of each style variable from -2 to 2 with an equal margin, rather than sampling from the prior \textit{Gaussian} distribution. We also show the generated samples by changing the random noise $z$ from -2 to 2 in Figure \ref{Fig.sub1.2-2}. In Figure \ref{Fig.sub1.1-1}, a small value of $h_{1}$ denotes a narrower digit whereas a high value corresponds to a wide digit. In Figure \ref{Fig.sub1.1-2}, $h_{2}$ smoothly controls the rotation. In Figure \ref{Fig.sub1.2-1}, we learn a special attribute, i.e., the center of gravity. A larger $h_{3}$ value means a higher center of gravity. In Figure \ref{Fig.sub1.2-2}, some digits have special fonts, such as $7$ with a bar in the middle and $8$ with an open loop in the top, which means  our Dual-AAE treats the digital special font as random noise.

\section{Conclusion}

This paper introduces a novel generative clustering algorithm, Dual Adversarial Auto-Encoder (Dual-AAE). It simultaneously maximizes the likelihood function and mutual information, and derives a new evidence lower bound by variational inference. Dual-AAE also introduces a new clustering regularization term to replace the adversarial training manner of the category variable, which eases the algorithm training procedure. Dual-AAE is optimized by maximizing the empirical estimator of the evidence lower bound. We evaluate our model on four benchmark databases, and the results show that it can outperform state-of-the-art clustering methods. Since Dual-AAE can learn disentangled representations better, we plan to extend this algorithm to semi-supervised learning and image style transfer in the future.

\bibliographystyle{IEEEtran}
\bibliography{DAAE}

\newpage

\appendices

\section{More Parameter Settings}\label{sect:appendix-network}

We use two regularization methods which are commonly used in the deep generative models. First, in the encoder and decoder networks, we use the batch normalization\cite{ioffe2015batch} in all layers except for the last layer. Additionally, we apply dropout \cite{srivastava2014dropout} noise to the hidden layers of the discriminator, which helps to increase the generalization ability of the discriminator. All loss functions are minimized by using the Adam \cite{kingma2014adam} algorithm.

{\bfseries MNIST}: In the latent variable layer, we set the dimensions of the category variable, the style variable, and the random noise to 10, 3 and 1, respectively. The architectures of encoder and decoder is as follows:
\begin{equation*}
\begin{split}
Input\rightarrow C(64,4,stride=2)&\rightarrow C(128,4,stride=2)\\
\rightarrow C(1024,7)\rightarrow C(128,1)\rightarrow FC&(10+4)\rightarrow DC(1024,1)\\
\rightarrow DC(128,7)\rightarrow DC(&64,4,stride=2)\\
\rightarrow DC(1,4,stride=2)\rightarrow si&gmoid()\rightarrow Output,
\end{split}
\end{equation*}
where $C(n,k)$ denotes a convolutional layer which has $n$ kernels of size $k\times k$ and stride 1. $FC(n)$ denotes a fully connected layer with $n$ output units. $DC(n,k)$ is a deconvolutional \cite{dosovitskiy2015learning} layer with $n$ kernels of size $k\times k$ and stride 1. In the encoder, leaky rectified linear units (LeakyReLU) \cite{maas2013rectifier} activation function is used for all layers except for the latent variable layer, and the slope of the leak is set to 0.1. In the decoder network, rectified linear units (ReLU) \cite{glorot2011deep} activation function is used for all layers except for the last layer. We use the same discriminator in all the experiments, of which the architecture is given as follows:
\begin{equation*}
\begin{split}
&Input\rightarrow FC(100)\rightarrow Drop(0.2)\rightarrow Relu()\\
\rightarrow FC(&100)\rightarrow Drop(0.2)\rightarrow Relu()\rightarrow FC(1)\rightarrow Output,
\end{split}
\end{equation*}
where $Drop(0.2)$ stands for a dropout regularization with a parameter 0.2. We use a stacked Auto-Encoder to initialize the encoder and the decoder.

{\bfseries HHAR}: In the latent variable layer, we set the dimensions of the category variable, the style variable, and the random noise to 6, 4 and 2, respectively. All layers are fully connected layer, and the architectures of encoder and decoder are 561-1000-1000-500-500-12 and 12-500-500-1000-1000-561. LeakyReLU activation function is used for all layers except for the last layer of encoder and decoder, and the slope of the leak is set to 0.1. All parameters in the network are initialized by a \textit{Gaussian} distribution $\mathcal{N}(0, 0.02)$.

{\bfseries STL-10}: In the latent variable layer, we set the dimensions of the category variable, the style variable and the random noise to 10, 4 and 4, respectively. The architectures of encoder and decoder are 2048-500-500-500-18 and 18-2000-500-1000-2048. The LeakyReLU activation function is also used, and the network parameters are initialized by a \textit{Gaussian} distribution $\mathcal{N}(0, 0.02)$.

\end{document}